\documentclass{article}
\usepackage{spconf,amsmath,graphicx,xcolor,cite,url}
\usepackage{fancyhdr}
\fancypagestyle{firstpage}
{
\fancyhf{}
\chead{\emph{Submitted to IEEE International Conference on Acoustics, Speech and Signal Processing (ICASSP), 2018.}}
}

\title{BUILDING COMPETITIVE DIRECT ACOUSTICS-TO-WORD MODELS\\FOR ENGLISH CONVERSATIONAL SPEECH RECOGNITION}
\name{Kartik Audhkhasi, Brian Kingsbury, Bhuvana Ramabhadran, George Saon, Michael Picheny\thanks{The authors thank Ewout Van Den Berg and Andrew Rosenberg of IBM, and Florian Metze of Carnegie Mellon University for useful discussions.}}
\address{IBM T. J. Watson Research Center, Yorktown Heights, New York}

\begin{document}
\ninept
\maketitle
\thispagestyle{firstpage}
\begin{abstract}
Direct acoustics-to-word (A2W) models in the end-to-end paradigm have received increasing attention compared to conventional sub-word based automatic speech recognition models using phones, characters, or context-dependent hidden Markov model states.
This is because A2W models recognize words from speech  
without any decoder, pronunciation lexicon, or externally-trained language model, making training and 
decoding with such models simple. Prior work has shown that A2W models require orders of magnitude more training data in order to perform comparably to conventional models. Our work also showed this accuracy gap when using the English Switchboard-Fisher data set. This paper describes a recipe to train an A2W model that closes this gap and is at-par with state-of-the-art sub-word based models. 
We achieve a word error rate of 8.8\%/13.9\% on the Hub5-2000 Switchboard/CallHome test sets without any decoder or language model. 
We find that model initialization, training data order, and regularization have the most impact on the A2W model performance. Next, we present 
a joint word-character A2W model that learns to first spell the word and then recognize it. This model  provides a rich output to the user instead of simple word hypotheses, making it especially useful in the case of words unseen or rarely-seen during training. 
\end{abstract}
\section{Introduction}
\label{sec:intro}
Conventional sub-word based automatic speech recognition (ASR) typically involves three models - acoustic model (AM), pronunciation model and decision tree (PMT), and 
language model (LM)~\cite{jelinek1997statistical}. The AM computes the probability $p_\text{AM}(\mathbf{A}|\mathbf{s})$ of the acoustics 
$\mathbf{A}$ given the sub-word units $\mathbf{s}$. The PMT models the probability $p_\text{PMT}(\mathbf{w}|\mathbf{s})$ of the word sequence 
$\mathbf{w}$ given the sub-word unit sequence $\mathbf{s}$. The LM acts as a prior $p_\text{LM}(\mathbf{w})$ on the
word sequence $\mathbf{w}$. Hence, finding the most likely sequence of words given the acoustics $\mathbf{A}$ becomes
a maximum aposteriori optimization problem over the following probability density function:
\begin{align}
     p_\text{ASR}(\mathbf{w}|\mathbf{A}) &\propto p_\text{LM}(\mathbf{w}) p(\mathbf{A}|\mathbf{w}) \\
                 &= p_\text{LM}(\mathbf{w}) \sum_{\mathbf{s}} p_\text{AM}(\mathbf{A}|\mathbf{s},\mathbf{w}) p_\text{PMT}(\mathbf{s}|\mathbf{w}) \\
                 &\approx p_\text{LM}(\mathbf{w}) \sum_{\mathbf{s}} p_\text{AM}(\mathbf{A}|\mathbf{s}) p_\text{PMT}(\mathbf{s}|\mathbf{w}) \;.
\end{align}
While both the AM and LM in modern ASR systems use deep neural networks (DNNs) and their variants~\cite{hinton2012deep}, the PMT is
usually based on decision trees and finite state transducers. 
Training the AM requires alignments between the acoustics and sub-word units, and several
iterations of model training and re-alignment. Recent end-to-end (E2E) models have obviated the need for 
aligning the sub-word units to the acoustics. Popular E2E models include recurrent neural networks (RNNs) 
trained with the connectionist temporal classification (CTC) loss function~\cite{graves2006connectionist, graves2014towards, zweig2016advances, sak2015fast, miao2015eesen, miao2016empirical, hannun2014deep, amodei2016deep} and the attention-based encoder-decoder RNNs~\cite{bahdanau2016end, bahdanau2014neural, maas2015lexicon, chan2016listen, chan2016latent}.These approaches are not truly E2E because they still use sub-word units, and hence require a decoder and separately-trained
LM to perform well.

In contrast, the recently-proposed direct acoustics-to-word (A2W) models~\cite{soltau2016neural,audhkhasi2017direct} train a single RNN to directly optimize $p_\text{ASR}(\mathbf{w}|\mathbf{A})$. 
This eliminates the need for sub-word units, pronunciation model, decision tree, decoder, and externally-trained LM, which significantly simplifies
the training and decoding process. However, prior research on A2W models has shown
that such models require several orders of magnitude more training data when compared with conventional
sub-word based models. This is because the A2W
models need sufficient acoustic training examples per word to train well.
For example,~\cite{soltau2016neural} used more than 125,000 hours of speech to train an A2W model with a vocabulary of nearly 100,000 words that matched the performance of a state-of-the-art CD state-based CTC model. Our prior work~\cite{audhkhasi2017direct} explored A2W models for the well-known English Switchboard task and presented
a few initialization techniques to effectively train such models with only 2000 hours of data. However, we still observed a gap of around 3-4\% absolute in WER between the Switchboard phone CTC and the A2W models on the Hub5-2000 evaluation set.

This paper further improves the state-of-the-art in A2W models for English conversational speech recognition. We 
present a training recipe that achieves WER of 8.8\%/13.9\% on the Switchboard/CallHome subsets of the Hub5-2000 evaluation set, compared to our previous best result of 13.0\%/18.8\%~\cite{audhkhasi2017direct}. These new results are at par with several state-of-the-art models that use
sub-word units, a decoder, and a LM. We quantify the gains made by each ingredient of our training recipe and conclude that model initialization, training data order, and regularization are the most important factors.

Next, we turn our attention to the issue of data sparsity while training A2W models. The conventional solution to this problem uses a sub-word unit-based model that
needs a decoder and LM during testing. As an alternative, we propose the spell and recognize (SAR) CTC model that
learns to first spell the word into its character sequence and then recognize it. Not only does this model retain
all advantages of a direct A2W model, it also provides rich hypotheses to the user which are readable
especially in the case of unseen or rarely-seen words. We illustrate the benefits of this model for out-of-vocabulary (OOV) words.

The next section discusses the baseline A2W model~\cite{audhkhasi2017direct}. Section~\ref{sec:recipe} discusses the proposed training recipe, an analysis of the impact of individual ingredients, and the results. Section~\ref{sec:sar}
presents our SAR model that jointly models words and characters. The paper concludes in Section~\ref{sec:concl} with a summary of findings.

\section{Baseline Acoustics-to-Word Model}\label{sec:baseline}

\subsection{CTC Loss}\label{sec:ctc_loss}
Conventional losses used for training neural networks, such as cross-entropy, require a one-to-one mapping (or alignment)
between the rows of the $T\times D$ input feature vector matrix $\mathbf{A}$ and length-$L$ output label sequence $\mathbf{y}$. The connectionist temporal
classification (CTC) loss relaxes this requirement by considering all possible alignments. It introduces a special \emph{blank symbol} $\phi$ that expands the length-$L$ target label sequence $\mathbf{y}$
to multiple length-$T$ label sequences $\tilde{\mathbf{y}}$ containing $\phi$, such that $\tilde{\mathbf{y}}$ maps
to $\mathbf{y}$ after removal of all repeating symbols and $\phi$. The CTC loss is then
\begin{align}
	p(\mathbf{y}|\mathbf{A}) &= \sum_{\tilde{\mathbf{y}}} p(\tilde{\mathbf{y}}|\mathbf{A}) = \sum_{\tilde{y}\in\Omega(\mathbf{y})} \prod_{t=1}^T p(\tilde{y}_t|\mathbf{a}_t) \;,
\end{align}
where the set $\Omega(\mathbf{y})$, is the set of CTC paths for $\mathbf{y}$, and $\mathbf{a}_t$/$y_t$ denote the $t^\text{th}$ elements of the sequences. 
A forward-backward algorithm efficiently computes the above loss function and its
gradient, which is then back-propagated through the neural network~\cite{graves2006connectionist}. The next section describes the baseline A2W model~\cite{audhkhasi2017direct}.

\subsection{Baseline A2W Model}\label{sec:base_a2w}
We used two standard training data sets for our experiments. The ``300-hour'' set contained 262 hours of segmented speech
from the Switchboard-1 audio with transcripts provided by Mississippi State University. The ``2000-hour'' set contained
an additional 1698 hours from the Fisher data collection and 15 hours from CallHome audio. 

We extracted 40-dimensional logMel filterbank features over 25 ms frames every 10 ms from the input speech signal.
We used stacking+decimation~\cite{sak2015fast}, where we stacked two successive frames and dropped every alternate frame
during training. This resulted in a stream of 80-dimensional acoustic feature vectors at half the frame rate of the original stream. The baseline models also used 100-dimensional i-vectors~\cite{saon2017english} for each speaker, resulting in 180-dimensional
acoustic feature vectors.

The baseline A2W model consisted of a 5-layer bidirectional LSTM (BLSTM) RNN with a 180-dimensional input and 320-dimensional hidden layers in the forward and backward directions. We picked words with at least 5 occurrences in the
training data in the vocabulary. This resulted in a 10,000-word output layer for the 300-hour A2W system and a 25,000-output layer for the 2000-hour system. As noted in~\cite{audhkhasi2017direct}, initialization is crucial to training
an A2W model. Thus, we initialized the A2W BLSTM with the BLSTM from a trained phone CTC model, and the final linear
layer using word embeddings trained using GloVe~\cite{pennington2014glove}.

Table~\ref{tab:base_results} gives the WERs of the baseline A2W and phone CTC models reported in~\cite{audhkhasi2017direct} on the Hub5-2000 Switchboard (SWB) and 
CallHome (CH) test sets. We performed the decoding of the A2W models via simple peak-picking over the output word posterior distribution, and removing repetitions and blank symbols. The phone CTC model used a full decoding graph and a LM. We observe that the 2000-hour A2W model lags behind the phone CTC model by 3.4\%/2.8\% absolute WER on SWB/CH, and the gap is much bigger for the 300-hour models. We next discuss our new training recipe.

\begin{table}[ht]
\caption{This table shows the Switchboard/CallHome WERs for our baseline 300/2000-hour A2W and phone CTC models from~\cite{audhkhasi2017direct}.}
\begin{center}
\begin{tabular}{|c|c|c|c|c|} \hline
\emph{Data (hrs)} &\emph{AM} & \emph{LM} & \emph{SWB} & \emph{CH} \\ \hline\hline
300 & A2W & - & 20.8 & 30.4 \\ \hline
300 & Phone CTC & Small & 14.5 & 25.1 \\ \hline\hline
2000 & A2W & - & 13.0 & 18.8 \\ \hline
2000 & Phone CTC & Big &  9.6 & 16.0 \\ \hline
\end{tabular}
\end{center}
\label{tab:base_results}
\end{table}
\vspace{-20pt}

\section{Updated Training Recipe}\label{sec:recipe}
Our prior experience with training A2W models led us to conclude that model initialization and regularization
are important aspects of training such models. One key reason for this is the fact that A2W models attempt
to solve the difficult problem of directly recognizing words from acoustics with a single neural network.
Hence, our previously-proposed strategy of initializing the A2W BLSTM with the phone CTC BLSTM and the final
linear layer with word embeddings gave WER gains. In this work, we started by exploring several
other strategies in the same spirit. All our new experiments were conducted in PyTorch~\cite{pytorch} with the following changes compared to~\cite{audhkhasi2017direct}:
\begin{itemize}
	\item We included delta and delta-delta coefficients because they slightly improved the WER. Hence, the total acoustic vector was 
of size 340 after stacking+decimation and appending the 100-dimensional i-vectors.
	\item We initialized all matrices to samples of a uniform distribution over $(-\epsilon,\epsilon)$
	where $\epsilon$ is the inverse of the $\sqrt{\text{fan-in}}$ or the size of the input vector~\cite{glorot2010understanding}. This takes the dimensionality of the input vectors at each layer in account, which improved convergence compared to our old
	strategy of using $\epsilon = 0.01$.
	\item In place of new-bob annealing, we kept a fixed learning rate for the first 10 epochs and decayed it by $\sqrt{0.5}$ every epoch.
\end{itemize}

\subsection{Training Data Order}\label{sec:data_order}
Training data order is an important consideration for sequence-to-sequence models such as E2E ASR systems
because such models operate on the entire input and output sequences. All training sequences have to be padded to the length of the longest sequence in the batch in order to do GPU tensor operations.
Random sequence order during batch creation is not memory-efficient because batches will contain a larger range of 
sequence lengths, which will lead to more wasteful padding on average. Hence, sequences have to be sorted
before batch creation. 

We compare the impact of sorting input acoustic sequences in order of ascending and descending length in
Table~\ref{tab:recipe_increment}. Our results show that ascending order gives significantly
better WER than sorting in descending order. The intuition behind this result is that 
shorter sequences are easier to train on initially, which enables the network to reach a better point in the parameter space.  This can be regarded as an instance of curriculum learning~\cite{Bengio2009}.

\begin{table}[ht]
\caption{This table shows the impact of various training strategies on the 300-hour A2W model.}
\begin{center}
\begin{tabular}{|l|c|c|} \hline
\emph{Model} & \emph{SWB} & \emph{CH} \\ \hline\hline
Descending Order & 23.0 & 30.7 \\ \hline
Ascending Order & 18.3 & 28.1 \\ \hline
+Momentum & 18.0 & 27.4 \\ \hline
+Dropout & 17.4 & 26.9 \\ \hline
+Output Projection & 16.9 & 26.3 \\ \hline
+Phone BLSTM Init. & 14.9 & 23.8 \\ \hline
{\bf +Bigger Model} & {\bf 14.6} & {\bf 23.6} \\ \hline\hline
Previous best~\cite{audhkhasi2017direct} & 20.8 & 30.4 \\ \hline
\end{tabular}
\end{center}
\label{tab:recipe_increment}
\end{table}
\vspace{-10pt}

\subsection{Momentum and Dropout}\label{sec:dropout}
We also experimented with Nesterov momentum-based stochastic gradient descent (SGD), which has been shown
to give better convergence compared to simple SGD on several tasks. We use the following parameter updates:
\begin{align}
	\mathbf{v}_n &= \rho \mathbf{v}_{n-1} + \lambda \nabla f(\Theta_{n-1}+\rho v_{n-1}) \\
	\Theta_{n} &= \Theta_{n-1} - \mathbf{v}_n \;,
\end{align}
where $\mathbf{v}_n$ is the \emph{velocity} or a running weighted-sum of the gradient of
the loss function $f$. The constant $\rho$ is usually set to $0.9$ and $\lambda$ is the learning rate, set to $0.01$ in our experiments. We also experimented with a dropout of 0.25 in order to prevent over-fitting. Table~\ref{tab:recipe_increment} shows that both momentum and dropout improve the WER.

\subsection{Output Projection Layer}\label{sec:proj}
In contrast with phone or character-based CTC models, A2W models have a large output size equal to the
size of the vocabulary. Prior research~\cite{sainath2013low} has shown that decomposing the output linear layer
of size $V \times D$ into two layers of sizes $V \times d$ and $d \times D$ with $d < D$ speeds-up
model training due to reduced number of parameters. We experimented with a projection layer of size 256 and found that it speeds-up training by a factor of 1.2x and also slightly improves the WER, which
we attribute to a reduction in over-fitting. 

\subsection{Phone BLSTM Initialization and Bigger Model}\label{sec:bigger}
We finally initialized our model with the phone CTC BLSTM which gave improvements in our previous work~\cite{audhkhasi2017direct}. As expected, this initialization lowered the WER, despite the presence of 
all the above useful strategies. With dropout in place, we trained a bigger 6-layer model with 512-dimensional
BLSTM and saw slight gains in WER.

\subsection{Final Model and Results}\label{sec:results}
We initialized the 2000-hour A2W model with the best 300-hour A2W model and used the same recipe for training. Table~\ref{tab:2k_wers} shows the WER of the resulting system along with our previous best A2W
WER and several other published results. We obtained a significant improvement of 4.2\%/4.9\% absolute
WER compared to our previous result. We also see that our direct A2W is at par with most hybrid CD
state-based and E2E models, while utilizing no decoder or LM. As noted in~\cite{saon2017english}, the CallHome test set
is more challenging than Switchboard because 36 out of 40 speakers in the latter appear in the
training set. The results on the CallHome
test set are especially good, where our A2W model matches the best result obtained using a
hybrid BLSTM~\cite{saon2017english} that used exactly the same acoustic features\footnotemark.\footnotetext{Adding additional FMLLR
features gives a WER of 7.2\%/12.7\%~\cite{saon2017english}.} 

\begin{table}[ht]
\caption{This table shows the WER of our current and previous-best A2W model trained on the 2000-hour Switchboard+Fisher set. We have also included several other published results for comparison. Results with $^*$ use data augmentation on the 2000-hour training set.}
\begin{center}
\begin{tabular}{|l|c|c|c|c|} \hline
\emph{Model} & \emph{Output} & \emph{Decoder/} & \emph{SWB} & \emph{CH} \\
& \emph{Units} & \emph{LM} & & \\ \hline\hline
BLSTM+LF MMI~\cite{povey2016purely} & CD state & Y/Y & 8.5 & 15.3 \\ \hline
LACE+LF MMI~\cite{xiong2016achieving} & CD state & Y/Y & 8.3 & 14.8 \\ \hline
Dilated Conv.~\cite{sercu2016dense} & CD state & Y/Y & 7.7 & 14.5 \\ \hline
BLSTM~\cite{saon2017english} & CD state & Y/Y & 7.7 & 13.9 \\ \hline
Iterated CTC~\cite{xiong2016microsoft} & Char & Y/Y & 11.3 & 18.7 \\ \hline
CTC$^*$~\cite{liu2017gram} & Char & Y/Y & 9.0 & 17.7 \\ \hline
Gram-CTC$^*$~\cite{liu2017gram} & Char N-gm & Y/Y & 7.9 & 15.8 \\ \hline
CTC+Gram-CTC$^*$~\cite{liu2017gram} & Char N-gm & Y/Y & 7.3 & 14.7 \\ \hline
RNN-T$^*$~\cite{battenberg2017exploring} & Char & Y/N & 8.5 & 16.4 \\ \hline
Attention$^*$~\cite{battenberg2017exploring} & Char & Y/N & 8.6 & 17.8 \\ \hline\hline
CTC A2W~\cite{audhkhasi2017direct} & Word & N/N & 13.0 & 18.8 \\ \hline
{\bf CTC A2W (current)} & {\bf Word} & {\bf N/N} & {\bf 8.8} & {\bf 13.9} \\ \hline
\end{tabular}
\end{center}
\label{tab:2k_wers}
\end{table}

We also note that our A2W model uses a 
vocabulary of 25,000 words which has an OOV rate of 0.5\%/0.8\% on the SWB/CH test sets. All the other
models used much bigger vocabularies and hence did not suffer from OOV-induced errors.

\subsection{Ablation Study}\label{sec:ablation}
\begin{figure}[htbp]
\begin{center}
\includegraphics[width=0.4\textwidth]{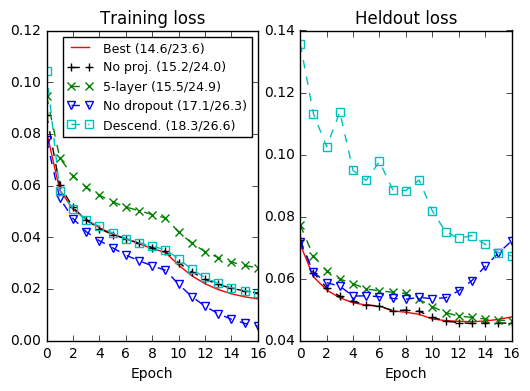}
\caption{This figure shows the results of the ablation experiment on our best 300-hour A2W model. The numbers in parentheses in the legend give the SWB/CH WERs.}
\label{fig:ablation_results}
\end{center}
\end{figure}

In order to understand the impact of individual components of our recipe, we conducted an ablation study
on our best 300-hour A2W model. We removed each component of the recipe while keeping others fixed, trained 
a model, and decoded the test sets. Figure~\ref{fig:ablation_results} shows the results of this experiment. We observed that changing the training data order from ascending to descending order of length
resulted in the biggest drop in performance. The second biggest factor was dropout - excluding it leads to
over-training because the heldout loss rises after epoch 10. Choosing a smaller 5-layer model
instead of 6-layer led to the next largest drop in WER. Finally, as expected, excluding the projection layer
had the least impact on WER.

Despite strong results, the A2W model does not give any meaningful output to the user in case of OOV words
but simply emits an \emph{``UNK''} tag. This is not a big problem for the Switchboard task because the OOV rates for the SWB/CH test sets are 0.5\%/0.8\% on the 25,000 word vocabulary. But other tasks
might be affected by the limited vocabulary. As a solution, the next section discusses our joint word-character model that aims to provide the user with a richer output that is especially useful for unseen or rarely-seen words.

\vspace{-5pt}
\section{Spell and Recognize Model}\label{sec:sar}
The advantage of the A2W model is that it directly emits word hypotheses by forward-passing acoustic
features through a RNN without needing a decoder or externally-trained LM. However, its vocabulary is fixed and OOV words cannot be recognized
by the system. Furthermore, words infrequently seen in the training data are not recognized well by the 
network due to insufficient training examples. Prior approaches to dealing with the above limitations completely rely on sub-word units. This
includes work on character models~\cite{xiong2016microsoft, hannun2014deep} and N-grams~\cite{liu2017gram}, RNN-Transducer~\cite{battenberg2017exploring}, RNN-Aligner~\cite{sak2017recurrent}.

In contrast, our approach is to have the best of both worlds by combining the ease of decoding a A2W model
with the flexibility of recognizing unseen/rarely-seen words with a character-based model. One natural
candidate is a multi-task learning (MTL) model containing a shared lower network, and two output networks
corresponding to the two tasks - recognizing words and characters. However, such an MTL network is not
suitable for our purpose because the recognized word and character sequences for an input speech utterance
are not guaranteed to be synchronized in time. This is because the CTC loss 
does not impose time-alignment on the output sequence.

The proposed spell and recognize (SAR) model circumvents this alignment problem by presenting training
examples that contain both words and characters. This allows us to continue to leverage an A2W framework without resorting to more complex graph-based decoding methodologies employed in, for example, word-fragment based systems~\cite{siohan2005fast,bhuvana2009fast,rastrow2009towards}. Consider the output word sequence ``THE CAT IS BLACK''
The SAR model 
uses the following target sequence:

{\footnotesize
{\color{red}b-t h e-e} {\color{blue}THE} {\color{red}b-c a e-t} {\color{blue}CAT} {\color{red}b-i e-s} {\color{blue}IS} {\color{red}b-b l a c e-k} {\color{blue}BLACK}}

where lowercase alphabets are the character targets, and b-/e- denote special prefixes for word beginning and end.
Hence, the model is trained to first spell the word and then recognize it. In contrast with a MTL model, the 
SAR model has a single softmax over words+characters in the output layer.

\subsection{Choice of Character Set}
We experimented with two character sets for the SAR model. The first one is the simple character set consisting
of a total of 41 symbols - alphabets a-z, digits 0-9, whitespace \_, and other punctuations. The second
character set is the one used in~\cite{zweig2016advances}, and includes separate character variants depending on position
in a word - beginning, middle, and end. It also includes special symbols for repeated characters, e.g. 
a separate symbol for \emph{ll}. The intuition behind this character set is that its symbols capture
more context as compared to the simple set, and also disambiguate legitimate character repetitions from
double peaks emitted by the CTC model. We observed that the performance with the latter character set is slightly better than using simple characters. Hence, we present results only for this case.

\subsection{Experiments and Results}
We restricted ourselves to the 300-hour set for experiments on the SAR model because it uses a 10,000 word vocabulary, leading to a higher OOV
rate than the 2000-hour set and also contains several rare words. We trained a 6-layer BLSTM with joint
word and character targets after preparing the output training sequences as described in the previous section.
We initialized the SAR BLSTM using the A2W BLSTM. The training recipe
was the same as for the A2W model presented previously. The SAR model permits three decodes:
\begin{itemize}
	\item {\bf Word}: Use only word predictions, similar to the A2W model.
	\item {\bf Characters}: Use only character predictions, and combine them into words using the word-begin characters.
	\item {\bf Switched}: Use the character predictions up to the previous word when the model predicts an ``UNK'' symbol, and use the word prediction everywhere else.
\end{itemize}

\vspace{-10pt}
\begin{table}[ht]
\caption{This table shows performance of the 300-hour SAR model for three types of decodes.}
\begin{center}
\begin{tabular}{|l|c|c|} \hline
\emph{Decode} & \emph{SWB} & \emph{CH} \\ \hline\hline
Word & 14.5 & 23.9 \\ \hline
Characters & 18.9 & 30.9 \\ \hline
Switched & 14.4 & 24.0 \\ \hline
\end{tabular}
\end{center}
\label{tab:sar_results}
\end{table}
\vspace{-10pt}

We observe that the SAR model gives comparable performance to the baseline A2W model, but additionally gives
meaningful output for OOVs, as illustrated by the following test set examples:

\vspace{5pt}
{\scriptsize \noindent\fbox{%
    \parbox{0.45\textwidth}{%
        \emph{REF:} SUCH AS LIKE (\%HESITATION) THE {\bf MURDERING} OF A COP OR\\
        \emph{HYP: }{\color{red}b-s u c e-h} {\color{blue}SUCH} {\color{red}\_ b-a e-s} {\color{blue}AS} {\color{red}\_ b-l i k e-e} {\color{blue}LIKE} {\color{red}\_ b-t h e-e} {\color{blue}THE} {\color{red}\_ b-u e-u} {\color{blue}UH} {\color{red}\bf\_ b-m u r d e r i n e-g} {\color{blue}\bf UNK} {\color{red}\_ b-o e-f} {\color{blue}OF} {\color{red}\_ b-a} {\color{blue}A} {\color{red}\_ b-s o e-p} {\color{blue}COP} {\color{red}\_ b-o e-r} {\color{blue}OR}\\
        \noindent\rule{8cm}{0.4pt} \\
        \emph{REF:} THAT IS RIGHT WE ARE WE ARE {\bf FURTHERING} HIGHER\\
        \emph{HYP:} {\color{red}b-t h a e-s} {\color{blue}THAT'S} {\color{red}\_ b-r i g h e-t} {\color{blue} RIGHT} {\color{red}\_ b-r h} {\color{blue}I RIGHT} {\color{red}\_ b-w e ' r e-e} {\color{blue}WE'RE} {\color{red}\_ b-w e ' r e-e} {\color{blue}WE'RE} {\color{red}\bf\_ b-f u r t h e r i n e-g} {\bf\color{blue}UNK} {\color{red}\_ b-h i g h e e-r} {\color{blue}HIGHER}\\
        \noindent\rule{8cm}{0.4pt} \\
        \emph{REF:} BUT SOMETIMES LIKE I JUST HAD TO DO THIS SUMMARY OF THIS ONE YOU KNOW THESE {\bf SCHOLARLY JOURNALS} AND STUFF\\
        \emph{HYP:} {\color{red}b-b u e-t} {\color{blue}BUT} {\color{red}\_ b-s o m e t i m e e-s} {\color{blue}SOMETIMES} {\color{red}\_ b-l i k e-e} {\color{blue}LIKE} {\color{red}\_ b-i I b-j u s e-t} {\color{blue}JUST} {\color{red}\_ b-h a e-d} {\color{blue}HAD} {\color{red}\_ b-t e-o} {\color{blue}TO} {\color{red}\_ b-d e-o} {\color{blue}DO} {\color{red}\_ b-t h e-e} {\color{blue}THIS} {\color{red}\_ b-s u mm e r e-y} {\color{blue}SUMMARY} {\color{red}\_ b-t h i e-s} {\color{blue}THIS} {\color{red}\_ b-o n e-e} {\color{blue}ONE} {\color{red}\_ b-y o e-u} {\color{blue}YOU} {\color{red}\_ b-k n o e-w} {\color{blue}KNOW} {\color{red}\_ b-t h e s e-e} {\color{blue}THESE} {\color{red}\bf\_ b-c o l a r l e-y} {\color{blue}\bf UNK} {\color{red}\bf\_ b-j o u r n a l e-s} {\color{blue}\bf UNK} {\color{red}\_ b-a n e-d} {\color{blue}AND} {\color{red}\_ b-s t u e-2f} {\color{blue}STUFF} 
    }%
}
}
\vspace{5pt}

The words in bold are OOVs. We observe that the SAR model emits the UNK tag in these cases, but the characters preceding it contain the spelling of the word. In some cases, this spelling is incorrect, e.g. "SCHOLARLY $\rightarrow$ COLARLY", but still is more meaningful to the user than the UNK tag. Future research will try to fix these errors using data-driven methods.

\section{Conclusion}\label{sec:concl}
Conventional wisdom and prior research suggests that direct acoustic-to-word (A2W) models require orders
of magnitude more data than sub-word unit-based models to perform competitively. This paper presents a recipe
to train a A2W model on the 2000-hour Switchboard+Fisher data set that performs at-par with several state-of-the-art hybrid and end-to-end models using sub-word units. We conclude that data order, model initialization, and regularization are crucial to obtaining a competitive A2W model with a WER of 8.8\%/13.9\% on the Switchboard/CallHome subsets of the Hub5-2000 test set. Next, we present a spell and recognize (SAR) model that learns to first spell a word and then recognize it. The proposed SAR model gives a rich and readable output
to the user while maintaining the training/decoding simplicity and performance of a A2W model. We show
some examples illustrating the SAR model's benefit for utterances containing OOV words.   

\bibliographystyle{IEEEbib}
\bibliography{ctc-word-model-icassp2018}

\begin{thebibliography}{10}

\bibitem{jelinek1997statistical}
F.~Jelinek,
\newblock {\em Statistical methods for speech recognition},
\newblock MIT press, 1997.

\bibitem{hinton2012deep}
G.~Hinton, L.~Deng, D.~Yu, G.~E. Dahl, A.~Mohamed, N.~Jaitly, A.~Senior,
  V.~Vanhoucke, P.~Nguyen, T.~N. Sainath, and B.~Kingsbury,
\newblock ``{Deep neural networks for acoustic modeling in speech recognition:
  The shared views of four research groups},''
\newblock {\em IEEE Signal Processing Magazine}, vol. 29, no. 6, pp. 82--97,
  2012.

\bibitem{graves2006connectionist}
A.~Graves, S.~Fern{\'a}ndez, F.~Gomez, and J.~Schmidhuber,
\newblock ``Connectionist temporal classification: labelling unsegmented
  sequence data with recurrent neural networks,''
\newblock in {\em Proc. ICML}, 2006, pp. 369--376.

\bibitem{graves2014towards}
A.~Graves and N.~Jaitly,
\newblock ``Towards end-to-end speech recognition with recurrent neural
  networks.,''
\newblock in {\em ICML}, 2014, vol.~14, pp. 1764--1772.

\bibitem{zweig2016advances}
G.~Zweig, C.~Yu, J.~Droppo, and A.~Stolcke,
\newblock ``Advances in all-neural speech recognition,''
\newblock in {\em Proc. ICASSP}, 2017, pp. 4805--4809.

\bibitem{sak2015fast}
H.~Sak, A.~Senior, K.~Rao, and F.~Beaufays,
\newblock ``Fast and accurate recurrent neural network acoustic models for
  speech recognition,''
\newblock in {\em Proc. Interspeech}, 2015.

\bibitem{miao2015eesen}
Y.~Miao, M.~Gowayyed, and F.~Metze,
\newblock ``{EESEN: End-to-end speech recognition using deep RNN models and
  WFST-based decoding},''
\newblock in {\em Proc. ASRU}, 2015, pp. 167--174.

\bibitem{miao2016empirical}
Y.~Miao, M.~Gowayyed, X.~Na, T.~Ko, F.~Metze, and A.~Waibel,
\newblock ``{An empirical exploration of CTC acoustic models},''
\newblock in {\em Proc. ICASSP}, 2016, pp. 2623--2627.

\bibitem{hannun2014deep}
A.~Hannun, C.~Case, J.~Casper, B.~Catanzaro, G.~Diamos, E.~Elsen, R.~Prenger,
  S.~Satheesh, S.~Sengupta, A.~Coates, and A.~Y. Ng,
\newblock ``Deep speech: Scaling up end-to-end speech recognition,''
\newblock {\em arXiv preprint arXiv:1412.5567}, 2014.

\bibitem{amodei2016deep}
D.~Amodei et~al.,
\newblock ``{Deep speech 2: End-to-end speech recognition in English and
  Mandarin},''
\newblock in {\em International Conference on Machine Learning}, 2016, pp.
  173--182.

\bibitem{bahdanau2016end}
D.~Bahdanau, J.~Chorowski, D.~Serdyuk, P.~Brakel, and Y.~Bengio,
\newblock ``End-to-end attention-based large vocabulary speech recognition,''
\newblock in {\em Proc. ICASSP}, 2016, pp. 4945--4949.

\bibitem{bahdanau2014neural}
D.~Bahdanau, K.~Cho, and Y.~Bengio,
\newblock ``Neural machine translation by jointly learning to align and
  translate,''
\newblock in {\em Proc. ICLR}, 2015.

\bibitem{maas2015lexicon}
A.~L. Maas, Z.~Xie, D.~Jurafsky, and A.~Y. Ng,
\newblock ``Lexicon-free conversational speech recognition with neural
  networks,''
\newblock in {\em Proc. HLT-NAACL}, 2015, pp. 345--354.

\bibitem{chan2016listen}
W.~Chan, N.~Jaitly, Q.~Le, and O.~Vinyals,
\newblock ``Listen, attend and spell: A neural network for large vocabulary
  conversational speech recognition,''
\newblock in {\em Proc. ICASSP}, 2016, pp. 4960--4964.

\bibitem{chan2016latent}
W.~Chan, Y.~Zhang, Q.~Le, and N.~Jaitly,
\newblock ``Latent sequence decompositions,''
\newblock {\em arXiv preprint arXiv:1610.03035}, 2016.

\bibitem{soltau2016neural}
H.~Soltau, H.~Liao, and H.~Sak,
\newblock ``{Neural Speech Recognizer: Acoustic-to-Word LSTM Model for Large
  Vocabulary Speech Recognition},''
\newblock in {\em Proc. Interspeech}, 2017.

\bibitem{audhkhasi2017direct}
K.~Audhkhasi, B.~Ramabhadran, G.~Saon, M.~Picheny, and D.~Nahamoo,
\newblock ``{Direct acoustics-to-word models for English conversational speech
  recognition},''
\newblock in {\em Proc. Interspeech}, 2017, pp. 959--963.

\bibitem{saon2017english}
G.~Saon, G.~Kurata, T.~Sercu, K.~Audhkhasi, S.~Thomas, D.~Dimitriadis, X.~Cui,
  B.~Ramabhadran, M.~Picheny, L.~Lim, B.~Roomi, and P.~Hall,
\newblock ``English conversational telephone speech recognition by humans and
  machines,''
\newblock in {\em Proc. Interspeech}, 2017.

\bibitem{pennington2014glove}
J.~Pennington, R.~Socher, and C.~D. Manning,
\newblock ``{Glove: Global Vectors for Word Representation},''
\newblock in {\em Proc. EMNLP}, 2014, vol.~14, pp. 1532--1543.

\bibitem{pytorch}
``{PyTorch},'' \url{https://github.com/pytorch/pytorch}.

\bibitem{glorot2010understanding}
X.~Glorot and Y.~Bengio,
\newblock ``Understanding the difficulty of training deep feedforward neural
  networks,''
\newblock in {\em Proc. AISTATS}, 2010, pp. 249--256.

\bibitem{Bengio2009}
Y.~Bengio, J.~Louradour, R.~Collobert, and J.~Weston,
\newblock ``Curriculum learning,''
\newblock in {\em Proc. ICML}, 2009.

\bibitem{sainath2013low}
T.~N. Sainath, B.~Kingsbury, V.~Sindhwani, E.~Arisoy, and B.~Ramabhadran,
\newblock ``Low-rank matrix factorization for deep neural network training with
  high-dimensional output targets,''
\newblock in {\em Proc. ICASSP}, 2013, pp. 6655--6659.

\bibitem{povey2016purely}
D.~Povey, V.~Peddinti, D.~Galvez, P.~Ghahremani, V.~Manohar, X.~Na, Y.~Wang,
  and S.~Khudanpur,
\newblock ``{Purely Sequence-Trained Neural Networks for ASR Based on
  Lattice-Free MMI},''
\newblock in {\em Proc. Interspeech}, 2016, pp. 2751--2755.

\bibitem{xiong2016achieving}
W.~Xiong, J.~Droppo, X.~Huang, F.~Seide, M.~Seltzer, A.~Stolcke, D.~Yu, and
  G.~Zweig,
\newblock ``Achieving human parity in conversational speech recognition,''
\newblock {\em arXiv preprint arXiv:1610.05256}, 2016.

\bibitem{sercu2016dense}
T.~Sercu and V.~Goel,
\newblock ``Dense prediction on sequences with time-dilated convolutions for
  speech recognition,''
\newblock {\em arXiv preprint arXiv:1611.09288}, 2016.

\bibitem{xiong2016microsoft}
W.~Xiong, J.~Droppo, X.~Huang, F.~Seide, M.~Seltzer, A.~Stolcke, D.~Yu, and
  G.~Zweig,
\newblock ``{The Microsoft 2016 conversational speech recognition system},''
\newblock in {\em Proc. ICASSP}, 2017, pp. 5255--5259.

\bibitem{liu2017gram}
H.~Liu, Z.~Zhu, X.~Li, and S.~Satheesh,
\newblock ``{Gram-CTC: Automatic Unit Selection and Target Decomposition for
  Sequence Labelling},''
\newblock in {\em Proc. ICML}, 2017.

\bibitem{battenberg2017exploring}
E.~Battenberg, J.~Chen, R.~Child, A.~Coates, Y.~Gaur, Y.~Li, H.~Liu,
  S.~Satheesh, D.~Seetapun, A.~Sriram, and Z.~Zhu,
\newblock ``Exploring neural transducers for end-to-end speech recognition,''
\newblock {\em arXiv preprint arXiv:1707.07413}, 2017.

\bibitem{sak2017recurrent}
H.~Sak, M.~Shannon, K.~Rao, and F.~Beaufays,
\newblock ``{Recurrent Neural Aligner: An Encoder-Decoder Neural Network Model
  for Sequence to Sequence Mapping},''
\newblock {\em Proc. Interspeech 2017}, pp. 1298--1302, 2017.

\bibitem{siohan2005fast}
O.~Siohan and M.~Bacchiani,
\newblock ``Fast vocabulary-independent audio search using path-based graph
  indexing,''
\newblock in {\em Proc. European Conference on Speech Communication and
  Technology}, 2005.

\bibitem{bhuvana2009fast}
B.~Ramabhadran, A.~Sethy, J.~Mamou, B.~Kingsbury, and U.~Chaudhari,
\newblock ``Fast decoding for open vocabulary spoken term detection,''
\newblock in {\em Proc. NAACL-HLT}, 2009, pp. 277--280.

\bibitem{rastrow2009towards}
A.~Rastrow, A.~Sethy, B.~Ramabhadran, and F.~Jelinek,
\newblock ``{Towards using hybrid word and fragment units for vocabulary
  independent LVCSR systems},''
\newblock in {\em Interspeech}, 2009, pp. 1931--1934.

\end{thebibliography}

\end{document}